\documentclass[11pt,a4paper]{article}
\usepackage[hyperref]{acl2019}
\usepackage{times}
\usepackage{latexsym}
\usepackage{booktabs}
\usepackage{url}

\aclfinalcopy % Uncomment this line for the final submission
\def\aclpaperid{1420} %  Enter the acl Paper ID here

\usepackage{textcomp}
\usepackage{multirow}
\usepackage{xspace}
\usepackage{tikz}
\usepackage{pgf}
\usepackage{arydshln}

\newcommand{\totaldata}{77,563\@\xspace}
\newcommand{\LoweCites}{315\@\xspace}

\newcommand{\tightparagraph}[1]{\noindent\textbf{#1:}}

\usetikzlibrary{arrows.meta,backgrounds,positioning,fit,calc,intersections,shapes,patterns}
\tikzset{not allowed/.style={densely dotted, thick, color=red}}
\tikzset{required directed/.style={shorten >=1pt, >=Stealth, ->}}
\tikzset{required undirected/.style={}}

\title{A Large-Scale Corpus for Conversation Disentanglement}

\author{
\begin{tabular}{ccc}
  Jonathan K. Kummerfeld$^1$\thanks{\tt \hspace{1mm} jkummerf@umich.edu} &
    Sai R. Gouravajhala$^1$ &
    Joseph J. Peper$^1$ \\
    Vignesh Athreya$^1$ &
    Chulaka Gunasekara$^2$ &
    Jatin Ganhotra$^2$ \\
    Siva Sankalp Patel$^2$ &
    Lazaros Polymenakos$^2$ &
    Walter S. Lasecki$^1$ \\[3pt]
  \multicolumn{3}{c}{
    \phantom{.} \hfill {\normalfont Computer Science \& Engineering$^1$} \hfill
  {\normalfont T.J. Watson Research Center$^2$} \hfill \phantom{.}
  } \\
  \multicolumn{3}{c}{
    \phantom{.} \hfill {\normalfont University of Michigan} \hfill
    \hspace{1cm}{\normalfont IBM Research AI} \hfill \phantom{.}
  } \\
\end{tabular}
}

\date{}

\begin{document}
\maketitle

\begin{abstract}
    Disentangling conversations mixed together in a single stream of messages is a difficult task, made harder by the lack of large manually annotated datasets.
    We created a new dataset of \totaldata messages manually annotated with reply-structure graphs that both disentangle conversations and define internal conversation structure.
    Our dataset is 16 times larger than all previously released datasets combined, the first to include adjudication of annotation disagreements, and the first to include context.
    We use our data to re-examine prior work, in particular, finding that 80\% of conversations in a widely used dialogue corpus are either missing messages or contain extra messages.
    Our manually-annotated data presents an opportunity to develop robust data-driven methods for conversation disentanglement, which will help advance dialogue research.
\end{abstract}

\section{Introduction}

When a group of people communicate in a common channel there are often multiple conversations occurring concurrently.
Often there is no explicit structure identifying conversations or their structure, such as in Internet Relay Chat (IRC), Google Hangout, and comment sections on websites.
Even when structure is provided it often has limited depth, such as threads in Slack, which provide one layer of branching.
In all of these cases, conversations are \emph{entangled}: all messages appear together, with no indication of separate conversations.
Automatic disentanglement could be used to provide more interpretable results when searching over chat logs, and to help users understand what is happening when they join a channel.
Over a decade of research has considered conversation disentanglement \citep{Shen:2006}, but using datasets that are either small \citep[2,500 messages,][]{elsner:2008} or not released \citep{Adams:2008}.

We introduce a conversation disentanglement dataset of \totaldata messages of IRC manually annotated with reply-to relations between messages.\footnote{
  \url{https://jkk.name/irc-disentanglement}
}
Our data is sampled from a technical support channel at 173 points in time between 2004 and 2018, providing a diverse set of speakers and topics, while remaining in a single domain.
Our data is the first to include context, which differentiates messages that start a conversation from messages that are responding to an earlier point in time.
We are also the first to adjudicate disagreements in disentanglement annotations, producing higher quality development and test sets.
We also developed a simple model that is more effective than prior work, and showed that having diverse data makes it perform better and more consistently.

We also analyze prior disentanglement work.
In particular, a recent approach from \citet{Lowe:2015,Lowe:2017:DD}.
By applying disentanglement to an enormous log of IRC messages, they developed a resource that has been widely used (over \LoweCites citations), indicating the value of disentanglement in dialogue research.
However, they lacked annotated data to evaluate the conversations produced by their method.
We find that 20\% of the conversations are completely right or a prefix of a true conversation; 58\% are missing messages, 3\% contain messages from other conversations, and 19\% have both issues.
As a result, systems trained on the data will not be learning from accurate human-human dialogues.

\section{Task Definition}

We consider a shared channel in which a group of people are communicating by sending messages that are visible to everyone.
We label this data with a \textbf{graph} in which messages are nodes and edges indicate that one message is a response to another.
Each connected component is a \textbf{conversation}.

\begin{figure}
\centering
\begin{tikzpicture}
[ every node/.style={
    node distance=2ex,
    inner sep=0pt,
    text width=6.5cm,
    font=\scriptsize\tt
  },
]

  \node (m2)  at (0,0) [anchor=north west] {\textcolor{ForestGreen}{\textbf{[03:05]}} <\textcolor{ForestGreen}{\textbf{delire}}> \textcolor{ForestGreen}{\textbf{hehe yes. does Kubuntu have 'KPackage'?}}};
  \node (m3) [below=5mm of m2.west, anchor=north west] {\textcolor{ForestGreen}{\textbf{=== delire found that to be an excellent interface to the apt suite in another distribution.}}};
  \node (m4) [below=6mm of m3.west, anchor=north west] {=== E-bola [...@...]  has joined \#ubuntu};
  \node (m5) [below=3mm of m4.west, anchor=north west] {\textcolor{NavyBlue}{\textbf{[03:06]}} <\textcolor{NavyBlue}{\textbf{BurgerMann}}> \textcolor{NavyBlue}{\textbf{does anyone know a consoleprog that scales jpegs fast and efficient?.. this digital camera age kills me when I have to scale photos :s}}};
  \node (m6) [below=7mm of m5.west, anchor=north west] {\textcolor{ForestGreen}{\textbf{[03:06]}} <\textcolor{ForestGreen}{\textbf{Seveas}}> \textcolor{ForestGreen}{\textbf{delire, yes}}};
  \node (m7) [below=3mm of m6.west, anchor=north west] {\textcolor{NavyBlue}{\textbf{[03:06]}} <\textcolor{NavyBlue}{\textbf{Seveas}}> \textcolor{NavyBlue}{\textbf{BurgerMann, convert}}};
  \node (m8) [below=3mm of m7.west, anchor=north west] {\textcolor{NavyBlue}{\textbf{[03:06]}} <\textcolor{NavyBlue}{\textbf{Seveas}}> \textcolor{NavyBlue}{\textbf{part of imagemagick}}};
  \node (m9) [below=3mm of m8.west, anchor=north west] {=== E-bola [...@...] has left \#ubuntu []};
  \node (m10) [below=3mm of m9.west, anchor=north west] {\textcolor{NavyBlue}{\textbf{[03:06]}} <\textcolor{NavyBlue}{\textbf{delire}}> \textcolor{NavyBlue}{\textbf{BurgerMann: ImageMagick}}};
  \node (m11) [below=3mm of m10.west, anchor=north west] {\textcolor{NavyBlue}{\textbf{[03:06]}} <\textcolor{NavyBlue}{\textbf{Seveas}}> \textcolor{NavyBlue}{\textbf{BurgerMann, i used that to convert 100's of photos in one command}}};
  \node (m12) [below=4mm of m11.west, anchor=north west] {\textcolor{NavyBlue}{\textbf{[03:06]}} <\textcolor{NavyBlue}{\textbf{BurgerMann}}> \textcolor{NavyBlue}{\textbf{Oh... I'll have a look.. thx =)}}};

  \draw [out=270,in=130,dashed,color=ForestGreen] (-0.3,0) to (m2.west);
  \draw [out=230,in=130,dashed,color=ForestGreen] (m2.west) to (m3.west);
  \draw [out=230,in=130,dashed,color=ForestGreen] (m2.west) to (m6.west);
  
  \draw [out=200,in=160,color=NavyBlue] (m5.west) to (m7.west);
  \draw [out=200,in=160,color=NavyBlue] (m7.west) to (m8.west);
  \draw [out=200,in=160,color=NavyBlue] (m5.west) to (m10.west);
  \draw [out=200,in=160,color=NavyBlue] (m7.west) to (m11.west);
  \draw [out=200,in=160,color=NavyBlue] (m7.west) to (m12.west);
  \draw [out=200,in=160,color=NavyBlue] (m10.west) to (m12.west);
\end{tikzpicture}

  \caption{\label{fig:examples}
  \texttt{\#Ubuntu} IRC log sample, earliest message first.
  Curved lines are our \emph{graph} annotations of reply structure, which define two \emph{conversations} shown with blue solid edges and green dashed edges.
  }
\end{figure}

Figure~\ref{fig:examples} shows an example of two entangled conversations and their graph structure.
It includes a message that receives multiple responses, when multiple people independently help \texttt{BurgerMann}, and the inverse, when the last message responds to multiple messages.
We also see two of the users, \texttt{delire} and \texttt{Seveas}, simultaneously participating in two conversations.
This multi-conversation participation is common.

The example also shows two aspects of IRC we will refer to later.
\textbf{Directed} messages, an informal practice in which a participant is named in the message.
These cues are useful for understanding the discussion, but only around 48\% of messages have them.
\textbf{System} messages, which indicate actions like users entering the channel.
These all start with \texttt{===}, but not all messages starting with \texttt{===} are system messages, as shown by the second message in Figure~\ref{fig:examples}.

\section{Related Work}

\tightparagraph{IRC Disentanglement Data}
The most significant work on conversation disentanglement is a line of papers developing data and models for the \texttt{\#Linux} IRC channel \citep{elsner:2008,Elsner:WILP:2009,Elsner:2010,Elsner:ACL:2011}.
Until now, their dataset was the only publicly available set of messages with annotated conversations (partially re-annotated by \citet{Mehri:2017:IJCNLP} with reply-structure graphs), and has been used for training and evaluation in subsequent work \citep{Wang:2009,Mehri:2017:IJCNLP,Jiang:NAACL:2018}.

We are aware of three other IRC disentanglement datasets.
First, \citet{Adams:2008} studied disentanglement and topic identification, but did not release their data.
Second, \citet{Riou:2015} annotated conversations and discourse relations in the \texttt{\#Ubuntu-fr} channel (French Ubuntu support).
Third, \citet{Lowe:2015,Lowe:2017:DD} heuristically extracted conversations from the \texttt{\#Ubuntu} channel.\footnote{
  This channel was first proposed as a useful data source by \citet{uthus-aha:2013:IJCNLP,Uthus:2013:FLAIRS,Uthus:2013:AAAI}, who identified messages relevant to the Unity desktop environment, and whether questions can be answered by the channel bot alone.
}
Their work opened up a new research opportunity by providing 930,000 disentangled conversations, and has already been the basis of many papers (\LoweCites citations), particularly on developing dialogue agents.
This is far beyond the size of resources previously collected, even with crowdsourcing \citep{hcw:2013}.
Using our data we provide the first empirical evaluation of their method.

\tightparagraph{Other Disentanglement Data}
IRC is not the only form of synchronous group conversation online.
Other platforms with similar communication formats have been studied in settings such as
classes \citep{wang-rose:icwsm:2008,Dulceanu:2016},
support communities \citep{mayfield:2012:SIGDIAL},
and customer service \citep{Du:2017:AAAI}.
Unfortunately, only one of these resources \citep{Dulceanu:2016} is available, possibly due to privacy concerns.

Another stream of research has used user-provided structure to get
conversation labels \cite{Shen:2006,Domeniconi:2016}
and reply-to relations \cite{wang-rose:2010:NAACLHLT,Wang:2011:SIGIR,ICWSM112840,Balali:2013,Balali:2014,Chen:2017}.
By removing these labels and mixing conversations they create a disentanglement problem.
While convenient, this risks introducing a bias, as people write differently when explicit structure is defined, and only a few papers have released data \cite{ABBOTT16.1126,Zhang:2017,louis-cohen:2015:EMNLP}.

\begin{table*}
    \centering
    \setlength{\tabcolsep}{4pt}
    \begin{tabular}{lllrrrrrr}
        \toprule
        Data         &             &                &          &       &             & Authors &         & Anno. \\
        Available?  &   Dataset   &                & Messages & Parts & Part Length &  / part & Context & / msg \\
        \midrule
        \multirow{9}{*}{Yes}
          & \multirow{7}{*}{This work} & \multicolumn{1}{l|}{Pilot}              &  1,250 &  9 & 100--332 msg &  19-48 & 0-100 & 1-5 \\
          &            & \multicolumn{1}{l|}{}                   & 47,500 & 95 &      500 msg &  33-95 &  1000 &   1 \\[-3pt]
          &            & \multicolumn{1}{l|}{Train \hfill \raisebox{0.4mm}{\hspace{1.4mm}\mbox{------------}\hspace{-1.4mm}}}              &  1,000 & 10 &      100 msg &  20-43 &  1000 & 3+a \\[-3pt]
          &            &                    & 18,963 & 48 &        1  hr & 22-142 &  1000 &   1 \\
          &            & Dev                &  2,500 & 10 &      250 msg & 76-167 &  1000 & 2+a \\
          &            & Test               &  5,000 & 10 &      500 msg & 79-221 &  1000 & 3+a \\
          &            & Channel 2          &  2,600 &  1 &        5  hr &    387 &     0 & 2+a \\[4pt]
          & \multicolumn{2}{l}{\citet{elsner:2008}}                &  2,500 &  1 &        5  hr &    379 &     0 & 1-6 \\
          & \multicolumn{2}{l}{\citet{Mehri:2017:IJCNLP}}          &    530 &  1 &      1\textonehalf\xspace hr &     54 &     0 &   3 \\
        \midrule
        \multirow{2}{*}{Request} &    \multicolumn{2}{l}{\citet{Riou:2015}}                  &  1,429 &  2 &  12 / 60  hr &  21/70 &     0 & 2/1 \\
           & \multicolumn{2}{l}{\citet{Dulceanu:2016}}              &    843 &  3 &  \textonehalf\xspace--1\textonehalf\xspace hr &    8-9 &   n/a &   1 \\
        \midrule
        \multirow{5}{*}{No} &    \multicolumn{2}{l}{\citet{Shen:2006}}              &  1,645 & 16 &  35--381 msg &   6-68 &   n/a &   1 \\
           & \multicolumn{2}{l}{\citet{Adams:2008}}             & 19,925 & 38 &  67--831 msg &      ? &     0 &   3 \\
           & \multicolumn{2}{l}{\citet{wang-rose:icwsm:2008}}   & $~$337 & 28 &    2--70 msg &      ? &   n/a & 1-2 \\
           & \multicolumn{2}{l}{\citet{mayfield:2012:SIGDIAL}}  &      ? & 45 &        1  hr &    3-7 &   n/a &   1 \\
           & \multicolumn{2}{l}{\citet{Guo:2017:EDB}}           &  1,500 &  1 &       48  hr &      5 &   n/a &   2 \\
        \bottomrule
    \end{tabular}
    \caption{\label{tab:data}
    Annotated disentanglement dataset comparison.
    Our data is much larger than prior work, one of the only released sets, and the only one with context and adjudication.
    `+a' indicates there was an adjudication step to resolve disagreements.
    `?' indicates the value is not in the paper and the authors no longer have access to the data.
    }
\end{table*}

\tightparagraph{Models}
\citet{elsner:2008} explored various message-pair feature sets and linear classifiers, combined with local and global inference methods.
Their system is the only publicly released statistical model for disentanglement of chat conversation, but most of the other work cited above applied similar models.
We evaluate their model on both our data and our re-annotated version of their data.
Recent work has applied neural networks \citep{Mehri:2017:IJCNLP,Jiang:NAACL:2018}, with slight gains in performance.

\tightparagraph{Graph Structure}
Within a conversation, we define a graph of reply-to relations.
Almost all prior work with annotated graph structures has been for threaded web forums \cite{schuth2007extracting,kim-wang-baldwin:2010:CONLL,Wang:2011:EMNLP}, which do not exhibit the disentanglement problem we explore.
Studies that do consider graphs for disentanglement have used small datasets \citep{Dulceanu:2016,Mehri:2017:IJCNLP} that are not always released \citep{wang-rose:icwsm:2008,Guo:2017:EDB}.

\section{Data}

We introduce a manually annotated dataset of \totaldata messages: 74,963 from the \texttt{\#Ubuntu} IRC channel,\footnote{
  \url{https://irclogs.ubuntu.com/}
} and 2,600 messages from the \texttt{\#Linux} IRC channel.\footnote{
  From \citet{elsner:2008}, including the 100 messages they did not annotate.
}
Annotating the \texttt{\#Linux} data enables comparison with \citet{elsner:2008}, while the \texttt{\#Ubuntu} channel has over 34 million messages, making it an interesting large-scale resource for dialogue research.
It also allows us to evaluate \citet{Lowe:2015,Lowe:2017:DD}'s widely used heuristically disentangled conversations.

When choosing samples we had to strike a balance between the number of samples and the size of each one.
We sampled the training set in three ways:
(1) 95 uniform length samples,
(2) 10 smaller samples to check annotator agreement,
and (3) 48 time spans of one hour that are diverse in terms of the number of messages, the number of participants, and what percentage of messages are directed.
For additional details of the data selection process, see the supplementary material.

\subsection{Dataset Comparison}

Table~\ref{tab:data} presents properties of our data and prior work on disentanglement in real-time chat.

\tightparagraph{Availability}
Only one other dataset, annotated twice, has been publicly released, and two others were shared when we contacted the authors.

\tightparagraph{Scale}
Our dataset is 31 times larger than almost any other dataset, the exception being one that was not released.
As well as being larger, our data is also based on many different points in time.
This is crucial because a single sample presents a biased view of the task.
Having multiple samples also means our training and evaluation sets are from different points in time, preventing overfitting to specific users or topics of conversation.

\tightparagraph{Context}
We are the first to consider the fact that IRC data is sampled from a continuous stream and the context prior to the sample is important.
In prior work, a message with no antecedent could either be the start of a conversation or a response to a message that occurs prior to the sample.

\tightparagraph{Adjudication}
Our labeling method is similar to prior work, but we are the first to perform adjudication of annotations.
While some cases were ambiguous, often one option was clearly incorrect.
By performing adjudication we can reduce these errors, creating high quality sets.

\subsection{Methodology}

\tightparagraph{Guidelines}
We developed annotation guidelines through three rounds of pilot annotations in which annotators labeled a set of messages and discussed all disagreements.
We instructed annotators to link each message to the one or more messages it is a response to.
If a message started a new conversation it was linked to itself.
We also described a series of subtle cases, using one to three examples to tease out differences.
These included when a question is repeated, when a user responds multiple times, interjections, etc.
For our full guidelines, see the supplementary material.
All annotations were performed using SLATE \cite{acl19slate}, a custom-built tool with features designed specifically for this task.\footnote{\url{https://jkk.name/slate}}

\tightparagraph{Adjudication}
Table~\ref{tab:data} shows the number of annotators for each subset of our data.
For the development, test, out-of-domain data, and a small set of the training data, we labeled each sample multiple times and then resolved all disagreements in an adjudication step.
During adjudication, there was no indication of who had given which annotation, and there was the option to choose a different annotation entirely.
In order to maximize the volume annotated, we did not perform adjudication for most of the training data.
Also, the 18,924 training message set initially only had 100 messages of context per sample, and we later added another 900 lines and checked every message that was not a reply to see if it was a response to something in the additional context.

\begin{figure}
\centering
  \scriptsize
  \tt
  \frenchspacing
  \setlength{\tabcolsep}{3pt}

  \begin{tabular}{l p{7cm}}
       & [21:29] <MOUD> that reminds me... how can I use CTRL+C/V on terminal? \\[1pt]
       & [21:29] <MonkeyDust> MOUD ctrl ins pasts \\[1pt]
       & [21:29] <nacc> MOUD: it depends on your terminal application, in gnome-terminal ... \\[1pt]
    \textbf{->} & \textcolor{OliveGreen}{\textbf{[21:30]} <\textbf{MOUD}> \textbf{-.-}} \\[11pt]

       & [17:35] <Moae> i have to remove LCDproc ... \\[1pt]
       & [17:38] <Madsy> Moae: sudo make uninstall \&\& make clean? :-) \\[1pt]
       & [17:39] <Madsy> Open the makefile and see what the targets are. \\[1pt]
    \textbf{->} & \textcolor{OliveGreen}{\textbf{[17:40]} <\textbf{Madsy}> \textbf{Moae: Don't message people in private please. It's ...}} \\[1pt]
       & [17:42] <Moae> Madsy: sorry \\[1pt]
       & [17:42] <Moae> Madsy where i have to launch the command? \\[1pt]
  \end{tabular}

  \caption{\label{fig:ambiguity}
  Examples of annotation ambiguity.
  Top: The message from \texttt{MOUD} could be a response to either \texttt{nacc} or \texttt{MonkeyDust}.
  Bottom: The message from \texttt{Madsy} could be part of this conversation or a separate exchange between the same users.
  }
\end{figure}

\tightparagraph{Annotators}
The annotators were all fluent English speakers with a background in computer science (necessary to understand the technical content): a postdoc, a master's student, and three CS undergraduates.
All adjudication was performed by the postdoc, who is a native English speaker.

\tightparagraph{Time}
Annotations took between 7 and 11 seconds per message depending on the complexity of the discussion, and adjudication took 5 seconds per message.
Overall, we spent approximately 240 hours on annotation and 15 hours on adjudication.

\subsection{Annotation Quality} \label{sec:agreement}

Our annotations define two levels of structure:
(1) links between pairs of messages, and
(2) sets of messages, where each set is one conversation.
Annotators label (1), from which (2) can be inferred.
Table~\ref{tab:agreement} presents inter-annotator agreement measures for both cases.
These are measured in the standard manner, by comparing the labels from different annotators on the same data.
We also include measurements for annotations in prior work.

Figure~\ref{fig:ambiguity} shows ambiguous examples from our data to provide some intuition for the source of disagreements.
In both examples the disagreement involves one link, but the conversation structure in the second case is substantially changed.
Some disagreements in our data are mistakes, where one annotation is clearly incorrect, and some are ambiguous cases, such as these.
In Channel Two, we also see mistakes and ambiguous cases, including a particularly long discussion about a user's financial difficulties that could be divided in multiple ways (also noted by \citet{elsner:2008}).
% Note, does not include Elsner's subject 3 (unfamiliar with linux)
%%%Also note that unlike prior work we adjudicated all disagreements in these sets.

\tightparagraph{Graphs}
We measure agreement on the graph structure annotation using \citet{Cohen:1960}'s $\kappa$.
This measure of inter-rater reliability corrects for chance agreement, accounting for the class imbalance between linked and not-linked pairs.

Values are in the good agreement range proposed by \citet{altman:1990}, and slightly higher than for \citet{Mehri:2017:IJCNLP}'s annotations.
Results are not shown for \citet{elsner:2008} because they did not annotate graphs.

\tightparagraph{Conversations}
We consider three metrics:\footnote{
  Metrics such as Cohen's $\kappa$ and Krippendorff's $\alpha$ are not applicable to conversations because there is no clear mapping from one set of conversations to another.
}

(1) Variation of Information \citep[VI,][]{Meila:2007}.
A measure of information gained or lost when going from one clustering to another.
It is the sum of conditional entropies $H(Y | X) + H(X | Y)$, where $X$ and $Y$ are clusterings of the same set of items.
We consider a scaled version, using the bound for $n$ items that VI$(X ; Y) \le \log(n)$, and present $1 - $VI so that larger values are better.

(2) One-to-One Overlap \citep[1-1,][]{elsner:2008}.
Percentage overlap when conversations from two annotations are optimally paired up using the max-flow algorithm.
We follow \citet{Mehri:2017:IJCNLP} and keep system messages.

(3) Exact Match F$_1$.
Calculated using the number of perfectly matching conversations, excluding conversations with only one message (mostly system messages).
This is an extremely challenging metric.
We include it because it is easy to understand and it directly measures a desired value (perfectly extracted conversations).

Our scores are higher in 4 cases and lower in 5.
Interestingly, while $\kappa$ was higher for us than \citet{Mehri:2017:IJCNLP}, our scores for conversations are lower.
This is possible because a single link can merge two conversations, meaning a single disagreement in links can cause a major difference in conversations.
This may reflect the fact that our annotation guide was developed for the Ubuntu channel, which differs in conversation style from the Channel Two data.
Manually comparing the annotations, there was no clear differences in the types of disagreements.

\begin{table}
    \centering
    \setlength{\tabcolsep}{4pt}
    \begin{tabular}{llcrrr}
        \toprule
                                                               &  & Graph & \multicolumn{3}{c}{Conversation} \\
            \multicolumn{2}{l}{Data}                            & $\kappa$ &   VI &  1-1 & F$_1$ \\
        \midrule                                                                        
            \multicolumn{2}{l}{Train (subset)}                      & 0.71 & 94.2 & 85.0 & 52.5 \\
            \multicolumn{2}{l}{Dev}                                 & 0.72 & 94.0 & 83.8 & 42.9 \\
            \multicolumn{2}{l}{Test}                                & 0.74 & 95.0 & 83.8 & 49.5 \\
            \multicolumn{2}{l}{Channel Two}                           & 0.72 & 90.4 & 75.9 & 28.2 \\
        \midrule
            \multicolumn{6}{l}{Subparts of Channel Two} \\
            \cmidrule(l){2-6}
            \multirow{2}{*}{Pilot} & This work                            & 0.68 & 90.9 & 82.4 & 43.5 \\
                                   & Elsner (\citeyear{elsner:2008})      &    - & 94.2 & 90.0 & 40.7 \\
            \cmidrule(l){2-6}
            Dev                    & This work                            & 0.74 & 92.2 & 81.7 & 27.5 \\
            \cmidrule(l){2-6}
            \multirow{2}{*}{Mehri} & This work                            & 0.73 & 86.2 & 71.9 & 22.2 \\
                                   & Mehri (\citeyear{Mehri:2017:IJCNLP}) & 0.67 & 91.3 & 80.7 & 38.7 \\
            \cmidrule(l){2-6}
            \multirow{2}{*}{Test}  & This work                            & 0.73 & 84.3 & 66.5 & 23.8 \\
                                   & Elsner (\citeyear{elsner:2008})      &    - & 80.8 & 62.4 & 20.6 \\
        \bottomrule
    \end{tabular}
    \caption{\label{tab:agreement}
      Inter-annotator agreement for graphs ($\kappa$) and conversations (1-1, VI, F$_1$).
      Our annotations are comparable to prior work, and $\kappa$ is in the good agreement range proposed by \citet{altman:1990}.
      We also adjudicated all disagreements to improve quality.
    }
\end{table}

Agreement is lower on the Channel Two data, particularly on its test set.
From this we conclude that there is substantial variation in the difficulty of conversation disentanglement across datasets.\footnote{
  \citet{Riou:2015} also observe this, noting that their French IRC data is less entangled than Elsner's, making it possible to achieve an agreement level of $0.95$.
}

\section{Evaluating Disentanglement Quality}

In this section, we propose new simple disentanglement models that perform better than prior methods, and re-examine prior work.
The models we consider are:

\tightparagraph{Previous} Each message is linked to the most recent non-system message before it.

\tightparagraph{\citet{Lowe:2017:DD}} A heuristic based on time differences and identifying directed messages.

\tightparagraph{\citet{elsner:2008}} A linear pairwise scoring model in which each message is linked to the highest scoring previous message, or none if all scores are below zero.

\tightparagraph{Linear} Our linear ranking model that scores potential antecedents using a feature-based model based on properties such as time, directedness, word overlap, and context.

\tightparagraph{Feedforward (FF)} Our feedforward model with the same features as the linear model, plus a sentence embedding calculated using an average of vectors from GloVe \citep{glove}.

\tightparagraph{Union} Run 10 FF models trained with different random seeds and combine their output by keeping all edges predicted.

\tightparagraph{Vote} Run 10 FF models and combine output by keeping the edges they all agree on. Link messages with no agreed antecedent to themselves.

\tightparagraph{Intersect} Conversations that 10 FF models agree on, and other messages as singleton conversations.

For Channel Two we also compare to \citet{Wang:2009} and \citet{Mehri:2017:IJCNLP}, but their code was unavailable, preventing evaluation on our data.
We exclude \citet{Jiang:NAACL:2018} as they substantially modified the dataset.
For details of models, including hyperparameters tuned on the development set, see the supplementary material.

\subsection{Results}

\begin{table}
  \centering
    \setlength{\tabcolsep}{5pt}
  \begin{tabular}{l ccc}
    \toprule
    System                    &           P   &           R      &           F   \\
    \midrule
    Previous                  &         35.7*  &          34.4*  &          35.0*  \\
    Linear                    &         64.7\phantom{*} & 62.3\phantom{*} & 63.5\phantom{*}  \\
    Feedforward               &         73.7*  &          71.0*  &          72.3*   \\
    \phantom{a}x10 union      &         64.3\phantom{*} & \textbf{79.7}*  & 71.2*   \\
    \phantom{a}x10 vote       & \textbf{74.9}* &           72.2* &  \textbf{73.5}*  \\
    \bottomrule
  \end{tabular}
  \caption{\label{tab:res-graph}
  \emph{Graph} results on the Ubuntu test set.
  * indicates a significant difference at the 0.01 level compared to Linear.
  }
\end{table}

\tightparagraph{Graphs}
Table~\ref{tab:res-graph} presents precision, recall, and F-score over links.
Our models perform much better than the baseline.
As we would expect, vote has higher precision, while union has higher recall.
Vote has higher recall than a single feedforward model because it identifies more of the self-link cases (its default when there is no agreement).

\begin{table}
  \centering
    \setlength{\tabcolsep}{5pt}
  \begin{tabular}{l rrrrr}
    \toprule
    System                          & VI & 1-1 & P & R & F \\
    \midrule
    Previous                        & 66.1 & 27.6 &  0.0 &  0.0 &  0.0 \\
    Linear                          & 88.9 & 69.5 & 19.3 & 24.9 & 21.8 \\
    Feedforward                     & 91.3 & 75.6 & 34.6 & 38.0 & 36.2 \\
    \phantom{a}x10 union            & 86.2 & 62.5 & 40.4 & 28.5 & 33.4 \\
    \phantom{a}x10 vote             & \textbf{91.5} & \textbf{76.0} & 36.3 & \textbf{39.7} & \textbf{38.0} \\
    \phantom{a}x10 intersect        & 69.3 & 26.6 & \textbf{67.0} & 21.1 & 32.1 \\
    \addlinespace[5pt]
    Lowe (\citeyear{Lowe:2017:DD})  & 80.6 & 53.7 & 10.8 &  7.6 &  8.9 \\
    Elsner (\citeyear{elsner:2008}) & 82.1 & 51.4 & 12.1 & 21.5 & 15.5 \\
    \bottomrule
  \end{tabular}
  \caption{\label{tab:res-conv}
  \emph{Conversation} results on the Ubuntu test set.
  Our new model is substantially better than prior work.
  Significance is not measured as we are unaware of methods for set structured data.
  }
\end{table}

\begin{table}
  \centering
    \setlength{\tabcolsep}{5pt}
  \begin{tabular}{l rr}
    \toprule
    Training Condition & Graph-F               & Conv-F     \\
    \midrule
    Standard           & 72.3\phantom{*} (0.4) & 36.2 (1.7) \\
    No context         & 72.3\phantom{*} (0.2) & 37.6 (1.6) \\
    1k random msg      & 63.0*           (0.4) & 21.0 (2.3) \\
    2x 500 msg samples & 61.4*           (1.8) & 20.4 (3.2) \\
    \bottomrule
  \end{tabular}
  \caption{\label{tab:res-data}
  Performance with different training conditions on the Ubuntu test set.
  For Graph-F, * indicates a significant difference at the 0.01 level compared to Standard.
  Results are averages over 10 runs, varying the data and random seeds.
  The standard deviation is shown in parentheses.
  }
\end{table}

\tightparagraph{Conversations}
Table~\ref{tab:res-conv} presents results on the metrics defined in Section~\ref{sec:agreement}.
There are three regions of performance.
First, the baseline has consistently low scores since it forms a single conversation containing all messages.
Second, \citet{elsner:2008} and \citet{Lowe:2017:DD} perform similarly, with one doing better on VI and the other on 1-1, though \citet{elsner:2008} do consistently better across the exact conversation extraction metrics.
Third, our methods do best, with x10 vote best in all cases except precision, where the intersect approach is much better.

\tightparagraph{Dataset Variations}
Table~\ref{tab:res-data} shows results for the feedforward model with several modifications to the training set, designed to test corpus design decisions.
Removing context does not substantially impact results.
Decreasing the data size to match \citet{elsner:2008}'s training set leads to worse results, both if the sentences are from diverse contexts (3rd row), and if they are from just two contexts (bottom row).
We also see a substantial increase in the standard deviation when only two samples are used, indicating that performance is not robust when the data is not widely sampled.

\subsection{Channel Two Results}

For channel Two, we consider two annotations of the same underlying text: ours and \citet{elsner:2008}'s.
To compare with prior work, we use the metrics defined by \citet[Shen]{Shen:2006} and \citet[Loc]{elsner:2008}.\footnote{
Loc is a Rand index that only counts messages less than 3 apart.
Shen calculates the F-score for each gold-system conversation pair, finds the max for each gold conversation, and averages weighted by the size of the gold conversation (this allows a predicted conversation to match to zero, one, or multiple gold conversations).
Following \citet{Wang:2009} and \citet{Mehri:2017:IJCNLP}, we include system messages in evaluation.
We also checked our metric implementations by removing system messages and calculating results for \citet{elsner:2008}'s output.
}
We do not use these for our data as they have been superseded by more rigorously studied metrics (VI for Shen) or make strong assumptions about the data (Loc).
We do not evaluate on graphs because \citet{elsner:2008}'s annotations do not include them.
This also prevents us from training our method on their data.

\begin{table}
    \centering
    \small
    \setlength{\tabcolsep}{4pt}
    \begin{tabular}{lllrrr}
        \toprule
        Test & Train & System & 1-1 & Loc & Shen \\
        \midrule
        \multirow{7}{*}{Elsner}
          & Ch 2 (Elsner)       & Elsner (\citeyear{elsner:2008})      & \underline{53.1} & \underline{81.9} & \underline{55.1} \\ 
          & Ch 2 (Elsner)       & Wang (\citeyear{Wang:2009})          &            47.0  &            75.1  &            52.8  \\
          & Ch 2 (Ours)         & Elsner (\citeyear{elsner:2008})      &            51.1  &            78.0  &            53.9  \\ 
          & Ch 2 (Ours)         & Feedforward                          &            52.1  &            77.8  &            53.8  \\
        \cmidrule(l){2-6}
          & Multiple            & Mehri (\citeyear{Mehri:2017:IJCNLP}) &            55.2  &            78.6  &            56.6  \\
          & n/a                 & Lowe (\citeyear{Lowe:2017:DD})       &            45.1  &            73.8  &            51.8  \\
          & Ubuntu              & Feedforward                          &    \textbf{57.5} &    \textbf{82.0} &    \textbf{60.5} \\
        \midrule
        \multirow{5}{*}{Ours}
          & Ch 2 (Elsner)       & Elsner (\citeyear{elsner:2008})      &            54.0  & \underline{81.2} &            56.3  \\ 
          & Ch 2 (Ours)         & Elsner (\citeyear{elsner:2008})      & \underline{59.7} &            80.8  & \underline{63.0} \\
          & Ch 2 (Ours)         & Feedforward                          &            57.7  &            80.3  &            59.8  \\
        \cmidrule(l){2-6}
          & n/a                 & Lowe (\citeyear{Lowe:2017:DD})       &            43.4  &            67.9  &            50.7  \\
          & Ubuntu              & Feedforward                          &    \textbf{62.8} &    \textbf{84.3} &    \textbf{66.6} \\
        \bottomrule
    \end{tabular}
    \caption{\label{tab:res:elsner}
    Results for different annotations of Channel Two.
    The \textbf{best result} is bold, and the \underline{best result with only Channel Two data} is underlined.
    }
\end{table}

\tightparagraph{Model Comparison}
For Elsner's annotations (top section of Table~\ref{tab:res:elsner}), their approach remains the most effective with just Channel Two data.
However, training on our Ubuntu data, treating Channel Two as an out-of-domain sample, yields substantially higher performance on two metrics and comparable performance on the third.
On our annotations (bottom section), we see the same trend.
In both cases, the heuristic from \citet{Lowe:2015,Lowe:2017:DD} performs poorly.
We suspect our model trained only on Channel Two data is overfitting, as the graph F-score on the training data is 94, whereas on the Ubuntu data it is 80.

\tightparagraph{Data Comparison}
Comparing the same models in the top and bottom section, scores are consistently higher for our annotations, except for the \citet{Lowe:2015,Lowe:2017:DD} heuristic.
Comparing the annotations, we find that their annotators identified between 250 and 328 conversations (mean 281), while we identify 257.
Beyond this difference it is hard to identify consistent variations in the annotations.
Another difference is the nature of the evaluation.
On Elsner's data, evaluation is performed by measuring relative to each annotators labels and averaging the scores.
On our data, we adjudicated the annotations, providing a single gold standard.
Evaluating our Channel-Two-trained Feedforward model on our two pre-adjudication annotations and averaging scores, the results are lower by 3.1, 1.8, and 4.3 on 1-1, Loc and Shen respectively.
This suggests that our adjudication process removes annotator mistakes that introduce noise into the evaluation.

\subsection{Evaluating \citet{Lowe:2015,Lowe:2017:DD}}

The previous section showed that only 10.8\% of the conversations extracted by the heuristic in \citet{Lowe:2015,Lowe:2017:DD} are correct (P in Table~\ref{tab:res-conv}).
We focus on precision because the primary use of their method has been to extract conversations to train and test dialogue systems, which will be impacted by errors in the conversations.
Recall errors (measuring missed conversations) are not as serious a problem because the Ubuntu chat logs are so large that even with low recall a large number of conversations will still be extracted.

\begin{figure}
\centering
  \scriptsize
  \tt

  \setlength{\tabcolsep}{2pt}
  \begin{tabular}{l p{6.5cm}}
    \textcolor{blue}{\bf Missed} & \textcolor{blue}{\textbf{[02:06]} <\textbf{TheBuntu}> \textbf{in virtualbox... win7 in VM... i have an ntfs partition.. How do i access that partition in VM ?}} \\[2pt]
           & [02:06] <L1nuxRules> share it with the vm \\[2pt]
           & [02:08] <L1nuxRules> anywy this is ubuntu so windows \&> /duv/null \\[2pt]
           & [02:09] <L1nuxRules> dev* \\[2pt]
    \textcolor{red}{\bf Extra}  & \textcolor{red}{\textbf{[02:11]} <\textbf{L1nuxRules}> \textbf{it shouldnt unless theres depency issues}} \\[2pt]
           & [02:11] <TheBuntu> L1nuxRules: how do i share with the vm... i dont see VM in share \\[2pt]
    \textcolor{blue}{\bf Missed} & \textcolor{blue}{\textbf{[02:12]} <\textbf{L1nuxRules}> \textbf{buntu if its virtuasl box click on setttings > shared folders}} \\[2pt]
    \textcolor{blue}{\bf Missed} & \textcolor{blue}{\textbf{[02:13]} <\textbf{TheBuntu}> \textbf{ok}} \\
  \end{tabular}

  \caption{\label{fig:error}
  An example conversation extracted by the heuristic from \citet{Lowe:2015,Lowe:2017:DD} with the messages it misses and the one it incorrectly includes.
  }
\end{figure}

\tightparagraph{Additional Metrics}
First, we must check this is not an artifact of our test set.
On our development set, P, R, and F are slightly higher (11.6, 8.1 and 9.5), but VI and 1-1 are slightly lower (80.0 and 51.7).
We can also measure performance as the distribution of scores over all of the samples we annotated.
The average precision was 10, and varied from 0 to 50, with 19\% of cases at 0 and 95\% below 23.
To avoid the possibility that we made a mistake running their code, we also considered evaluating their released conversations.
On the data that overlapped with our annotations, the precision was 9\%.
These results indicate that the test set performance is not an aberration: the heuristic's results are consistently low, with only about 10\% of output conversations completely right.

\tightparagraph{Error Types}
Figure~\ref{fig:error} shows an example heuristic output with several types of errors.
The initial question was missed, as was the final resolution, and in the middle there is a message from a separate conversation.
67\% of conversations were a subset of a true conversation (ie., only missed messages), and 3\% were a superset of a true conversation (ie., only had extra messages).
The subset cases were missing 1-187 messages (missing 56\% of the conversation on average) and the superset cases had 1-3 extra messages (an extra 31\% of the conversation on average).
The first message is particularly important because it is usually the question being resolved.
In 47\% of cases the first message is not the true start of a conversation.

It is important to note that the dialogue task the conversations were intended for only uses a prefix of each conversation.
For this purpose, missing the end of a conversation is not a problem.
In 9\% of cases, the conversation is a true prefix of a gold conversation.
Combined with the exact match cases, that means 20\% of the conversations are accurate as used in the next utterance selection task.
A further 9\% of cases are a continuous chunk of a conversation, but missing one or more messages at the start.

\begin{figure}
  \centering

  \includegraphics[width=\linewidth, trim={0 0 0 0}]{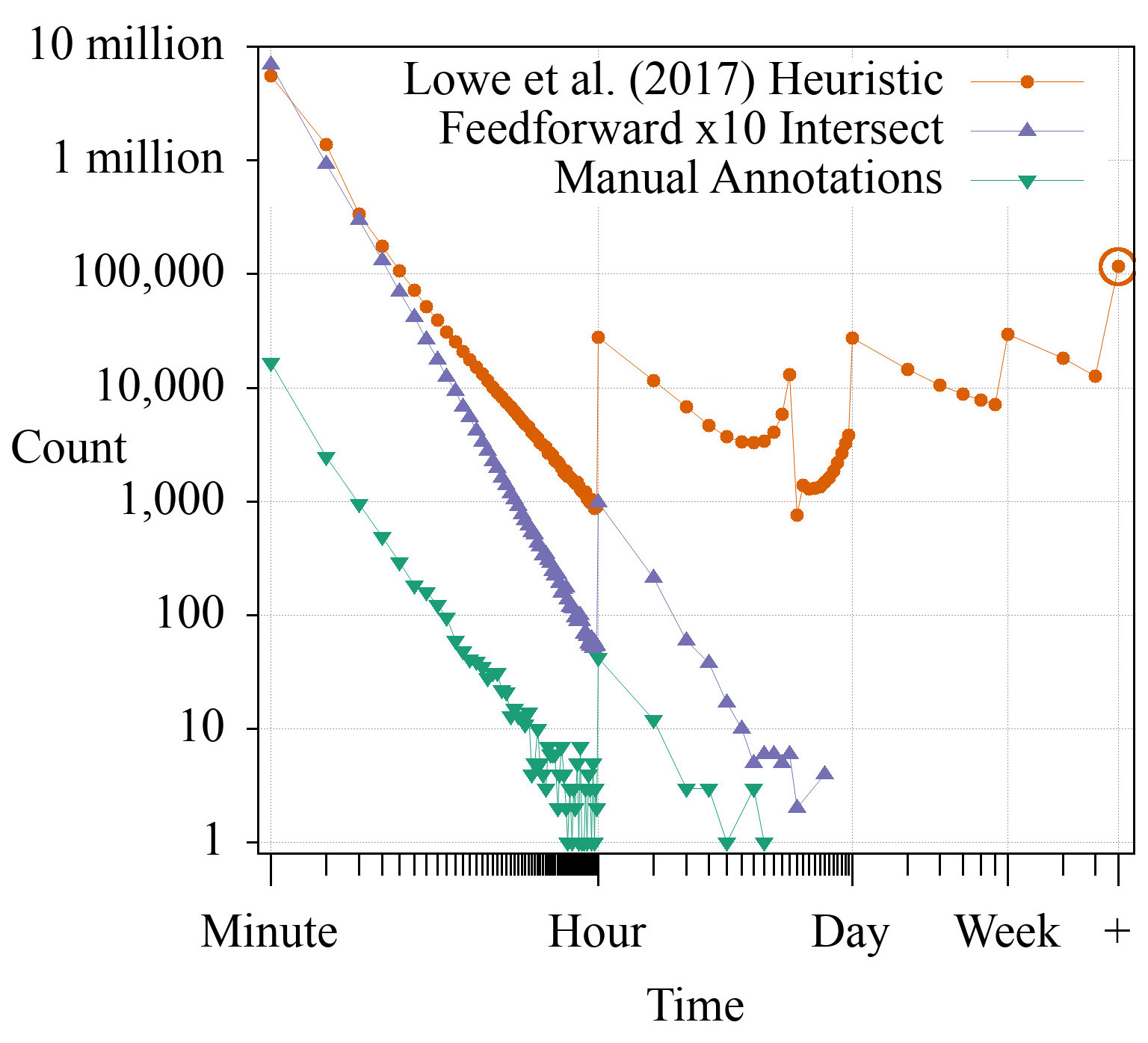}

  \caption{\label{fig:time-between-messages}
  Time between consecutive messages in conversations.
  Jumps are at points when the scale shifts as indicated on the x-axis.
  The circled upper right point is the sum over all larger values, indicating that messages weeks apart are often in the same conversation.
  }
\end{figure}

\tightparagraph{Long Distance Links}
One issue we observed is that conversations often spanned days.
We manually inspected a random sample: 20 conversations 12 to 24 hours long, and 20 longer than 24 hours.
All of the longer conversations and 17 of the shorter ones were clearly incorrect.\footnote{
The exceptions were two cases where a user thanked another user for their help the previous day, and one case where a user asked if another user ended up resolving their question.
}
This issue is not measured in the analysis above because our samples do not span days (they are 5.5 hours long on average when including context).
The original work notes this issue, but claims that it is rare.
We measured the time between consecutive messages in conversations and plot the frequency of each value in Figure~\ref{fig:time-between-messages}.\footnote{
  In 68,002 conversations there was a negative time difference because a message was out of order.
  To resolve this, we sorted the messages in each conversation by timestamp.
}
The figure indicates that the conversations often extend over days, or even more than a month apart (note the point in the top-right corner).
In contrast, our annotations rarely contain links beyond an hour, and the output of our model rarely contains links longer than 2 hours.

\tightparagraph{Causes}
To investigate possible reasons for these issues, we measured several properties of our data to test assumptions in the heuristic.
First, the heuristic assumes if all directed messages from a user are in one conversation, all undirected messages from the user are in the same conversation.
We find this is true 52.2\% of the time.
Second, it assumes that it is rare for two people to respond to an initial question.
In our data, of the messages that start a conversation and receive a response, 37.7\% receive multiple responses.
Third, that a directed message can start a conversation, which we find in 6.8\% of cases.
Fourth, that the first response to a question is within 3 minutes, which we find is true in 94.8\% of conversations.
Overall, these assumptions have mixed support from our data, which may be why the heuristic produces so few accurate conversations.

\begin{table}
  \centering
  \setlength{\tabcolsep}{5pt}
  \begin{tabular}{lllccc}
    \toprule
        Model  & Test & Train & MRR & R@1 & R@5 \\
    \midrule
        \multirow{4}{*}{DE}   & \multirow{2}{*}{Lowe} & Lowe & 0.75 & 0.61 & 0.94 \\
                              &                       & Ours & 0.63 & 0.45 & 0.90 \\
                              \cmidrule(l){3-6}
                              & \multirow{2}{*}{Ours} & Lowe & 0.72 & 0.57 & 0.93 \\
                              &                       & Ours & 0.76 & 0.63 & 0.94 \\
    \midrule
        \multirow{4}{*}{ESIM} & \multirow{2}{*}{Lowe} & Lowe & 0.82 & 0.72 & 0.97 \\
                              &                       & Ours & 0.69 & 0.53 & 0.92 \\
                              \cmidrule(l){3-6}
                              & \multirow{2}{*}{Ours} & Lowe & 0.78 & 0.67 & 0.95 \\
                              &                       & Ours & 0.83 & 0.74 & 0.97 \\
    \bottomrule
  \end{tabular}
  \caption{\label{tab:dialogue-results}
  Next utterance prediction results, with various models and training data variations.
  The decrease in performance when training on one set and testing on the other suggests they differ in content.
  }
\end{table}

\tightparagraph{Dialogue Modeling}
Most of the work building on \citet{Lowe:2017:DD} uses the conversations to train and evaluate dialogue systems.
To see the impact on downstream work, we constructed a next utterance selection task as described in their work, disentangling the entire \texttt{\#Ubuntu} logs with our feedforward model.
We tried two dialogue models: a dual-encoder \citep{Lowe:2017:DD}, and Enhanced Long Short-Term Memory \citep{Chen:ACL:2017}.
For full details of the task and model hyperparameters, see the supplementary material.

Table~\ref{tab:dialogue-results} show results when varying the training and test datasets.
Training and testing on the same dataset leads to higher performance than training on one and testing on the other.
This is true even though the heuristic data contains nine times as many training conversations.
This is evidence that our conversations are fundamentally different despite being derived from the same resource and filtered in the same way.
This indicates that our changes lead to quantitatively different downstream models.
Fortunately, the relative performance of the two models remains consistent across the two datasets.

\subsection{Re-Examining Disentanglement Research} \label{sec:analysis}

Using our data we also investigate other assumptions made in prior work.
The scale of our data provides a more robust test of these ideas.

\tightparagraph{Number of samples}
Table~\ref{tab:data} shows that all prior work with available data has considered a small number of samples.
In Table~\ref{tab:res-data}, we saw that training on less diverse data samples led to models that performed worse and with higher variance.
We can also investigate this by looking at performance on the different samples in our test set.
The difficulty of samples varies considerably, with the F-score of our model varying from 11 to 40 and annotator agreement scores before adjudication varying from 0.65 to 0.78.
The model performance and agreement levels are also strongly correlated, with a Spearman's rank correlation of 0.77.
This demonstrates the importance of evaluating on data from more than one point in time to get a robust estimate of performance.

\tightparagraph{How far apart consecutive messages in a conversation are}
\citet{elsner:2008} and \citet{Mehri:2017:IJCNLP} use a limit of 129 seconds,
\citet{Jiang:NAACL:2018} limit to within 1 hour,
\citet{Guo:2017:EDB} limit to within 8 messages,
and we limit to within 100 messages.
Figure~\ref{fig:time-between-messages} shows the distribution of time differences in our conversations.
94.9\% are within 2 minutes, and almost all are within an hour.
88.3\% are 8 messages or less apart, and 99.4\% are 100 or less apart.
This suggests that the lower limits in prior work are too low.
However, in Channel Two, 98\% of messages are within 2 minutes, suggesting this property is channel and sample dependent.

\tightparagraph{Concurrent conversations}
\citet{Adams:2008} forced annotators to label at most 3 conversations, while \citet{Jiang:NAACL:2018} remove conversations to ensure there are no more than 10 at once.
We find there are 3 or fewer 46.4\% of the time and 10 or fewer 97.3\% of the time (where time is in terms of messages, not minutes, and we ignore system messages),
Presumably the annotators in \citet{Adams:2008} would have proposed changes if the 3 conversation limit was problematic, suggesting that their data is less entangled than ours.

\tightparagraph{Conversation and message length}
\citet{Adams:2008} annotate blocks of 200 messages.
If such a limit applied to our data, 13.7\% of conversations would not finish before the cutoff point.
This suggests that their conversations are typically shorter, which is consistent with the previous conclusion that their conversations are less entangled.
\citet{Jiang:NAACL:2018} remove conversations with fewer than 10 messages, describing them as outliers, and remove messages shorter than 5 words, arguing that they were not part of real conversations.
Not counting conversations with only system messages, 83.4\% of our conversations have fewer than 10 messages, 40.8\% of which have multiple authors.
88.5\% of messages with less than 5 words are in conversations with more than one author.
These values suggest that these messages and conversations are real and not outliers.

\tightparagraph{Overall}
This analysis indicates that working from a small number of samples can lead to major bias in system design for disentanglement.
There is substantial variation across channels, and across time within a single channel.

\section{Conclusion}

Conversation disentanglement has been under-studied because of a lack of public, annotated datasets.
We introduce a new corpus that is larger and more diverse than any prior corpus, and the first to include context and adjudicated annotations.
Using our data, we perform the first empirical analysis of \citet{Lowe:2015,Lowe:2017:DD}'s widely used data, finding that only 20\% of the conversations their method produces are true prefixes of conversations.
The models we develop have already enabled new directions in dialogue research, providing disentangled conversations for DSTC 7 track 1 \citep{dstc19task1,ws18dstc} and will be used in DSTC 8.
We also show that diversity is particularly important for the development of robust models.
This work fills a key gap that has limited research, providing a new opportunity for understanding synchronous multi-party conversation online.

\section*{Acknowledgements}
We would like to thank Jacob Andreas, Greg Durrett, Will Radford, Ryan Lowe, and Glen Pink for helpful feedback on earlier drafts of this paper and the anonymous reviewers for their helpful suggestions.
This material is based in part on work supported by IBM as part of the Sapphire Project at the University of Michigan.
Any opinions, findings, conclusions or recommendations expressed above do not necessarily reflect the views of IBM.

\bibliography{acl19irc}
\bibliographystyle{acl_natbib}

\end{document}